\documentclass[conference]{IEEEtran}
\IEEEoverridecommandlockouts
\usepackage{graphicx} 
\usepackage[english]{babel}
\usepackage[utf8]{inputenc}
\usepackage[T1]{fontenc}
\usepackage{amsmath, amsthm}
\usepackage{amssymb}  
\usepackage{algorithm,algorithmicx}
\usepackage[noend]{algpseudocode}
\usepackage{mathtools}
\usepackage{helvet}
\usepackage{etoolbox}
\usepackage{titlesec}
\usepackage{caption}
\usepackage{booktabs}
\usepackage{xcolor} 
\usepackage{cite}
\usepackage{caption}
\usepackage{bbm}
\usepackage[numbers,sort]{natbib}
\algnewcommand{\algorithmicand}{\textbf{ and }}
\algnewcommand{\algorithmicor}{\textbf{ or }}
\algnewcommand{\OR}{\algorithmicor}
\algnewcommand{\AND}{\algorithmicand}
\algnewcommand{\var}{\texttt}
\usepackage{color}
\usepackage{xcolor}
\newcommand{\minisection}[1]{\vspace{0.025in} \noindent {\bf #1}\ }

\def\BibTeX{{\rm B\kern-.05em{\sc i\kern-.025em b}\kern-.08em
    T\kern-.1667em\lower.7ex\hbox{E}\kern-.125emX}}
\begin{document}

\title{Efficient Ground Vehicle\\ Path
Following in Game AI
}

\author{
Rodrigue de Schaetzen$^{1, 2}$,
Alessandro Sestini$^{3}$
\\
\textit{$^1$University of Waterloo, Canada}, 
\textit{$^2$Electronic Arts (EA)}, 
\textit{$^3$SEED - Electronic Arts (EA)}  \\
rdeschae@uwaterloo.ca, 
asestini@ea.com \\
}

\IEEEoverridecommandlockouts

\IEEEpubid{\makebox[\columnwidth]{979-8-3503-2277-4/23/\$31.00~\copyright2023 IEEE \hfill} 
\hspace{\columnsep}\makebox[\columnwidth]{ }}

\maketitle

\IEEEpubidadjcol

\begin{abstract}
This short paper presents an efficient path following solution for ground vehicles tailored to game AI. Our focus is on adapting established techniques to design simple solutions with parameters that are easily tunable for an efficient benchmark path follower. Our solution pays particular attention to computing a target speed which uses quadratic B\'ezier curves to estimate the path curvature. The performance of the proposed path follower is evaluated through a variety of test scenarios in a first-person shooter game, demonstrating its effectiveness and robustness in handling different types of paths and vehicles. We achieved a 70\% decrease in the total number of stuck events compared to an existing path following solution.
\end{abstract}


\section{Introduction and Related Work}
An important component of game AI is driving vehicles along predefined paths in an effective and efficient manner. The task of computing control actions that keep a vehicle along a path while tracking a target speed is referred to as the \emph{path following} problem. This problem has been extensively explored in robotics and control literature, particularly for applications in unmanned ground and aerial vehicles \cite{terlizzi2021vision, bacha2017}. In the context of game AI, path following presents particular challenges, including navigation in a wide range of environments, adhering to limited computational requirements, and achieving good performance across a range of vehicles for games that contain a wide selection of vehicle types. 

While many of the works from robotics and control literature can be applied to game AI, few papers discuss the various adaptations necessary to make use of these established techniques. Most of the works that consider path following for AI controlled vehicles are for car racing games \cite{onieva2012evolutionary, etlik2021fuzzy,melder2014racing}. \citet{onieva2012evolutionary} offer a comprehensive overview of their 7-part driving architecture, which utilizes several vehicle sensors to gather information about the vehicle's position relative to the track. The sensor readings along with a learned module determine the optimal parameter values in the different modules. Another notable related work is the fuzzy logic based self-driving racing car control system by \citet{etlik2021fuzzy}. The authors use a vision based lane detection system for online lane detection and fuzzy logic for generating position and velocity references. The work by \citet{melder2014racing} provide an introduction to Proportional-Integral-Derivative (PID) controllers for game AI, as well as brief descriptions of some of the PID variants. 

In this work, we consider the path following problem for controlling ground vehicles in game AI. We propose a general framework which may serve as a benchmark solution offering reasonable path following accuracy and efficiency for a wide range of vehicles and environments. We leverage established, effective, and computationally cheap techniques for speed and steering control and use an analytical result for computing the maximum curvature of a quadratic B\'ezier curve in a novel algorithm for computing the target speed. The performance of our approach is evaluated by running experiments in a first-person shooter game containing a suite of different vehicles. 

\section{Proposed Method}

\begin{figure}[!t]
    \centering
    \includegraphics[width=0.6\columnwidth]{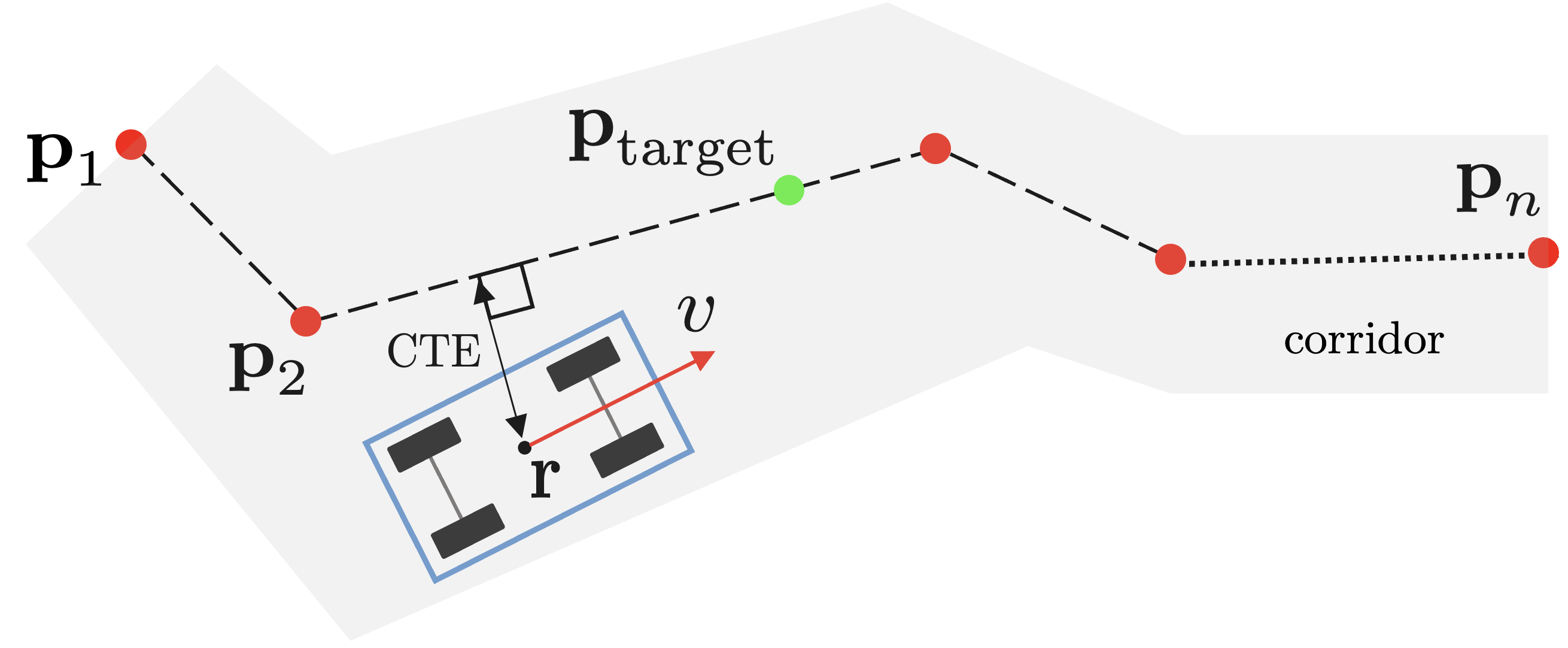}
    \caption{Ground vehicle (blue) following path (black) with corridor (grey) and waypoints (red).}
    \label{fig:prelim}
\end{figure}

\minisection{Problem Formulation.} We consider the path following problem of commanding a ground vehicle with arbitrary kinematics to move along a given path. The objective is to minimize the cross track error (CTE) between the vehicle's position $\textbf{r}\in\mathbb{R}^3$ (often defined at the centre of mass) and the path while maintaining a target speed $v_{\text{target}}$ and following a moving target point $\textbf{p}_{\text{target}}$ on the path. As a secondary objective, we wish to keep the vehicle within the safe boundaries of the path, commonly referred to as the path \emph{corridor}. The path $P$ is represented as a list of waypoints $(\textbf{p}_1, \textbf{p}_2, ..., \textbf{p}_n), \ \textbf{p}_i \in \mathbb{R}^3$ and may change significantly across frames. 
A sample path is shown in Figure~\ref{fig:prelim}.
We denote vehicle speed along the axis of the vehicle's forward vector as $v \in \mathbb{R}$. The sign of $v$ denotes the direction of the vehicle's motion, i.e., $v<0$ indicates the vehicle is going in reverse and vice versa. We assume there are two inputs to the game for changing the vehicle's motion: steering and throttle for lateral and longitudinal motion, respectively.

\begin{figure}[!t]
    \centering
    \includegraphics[scale=0.15]{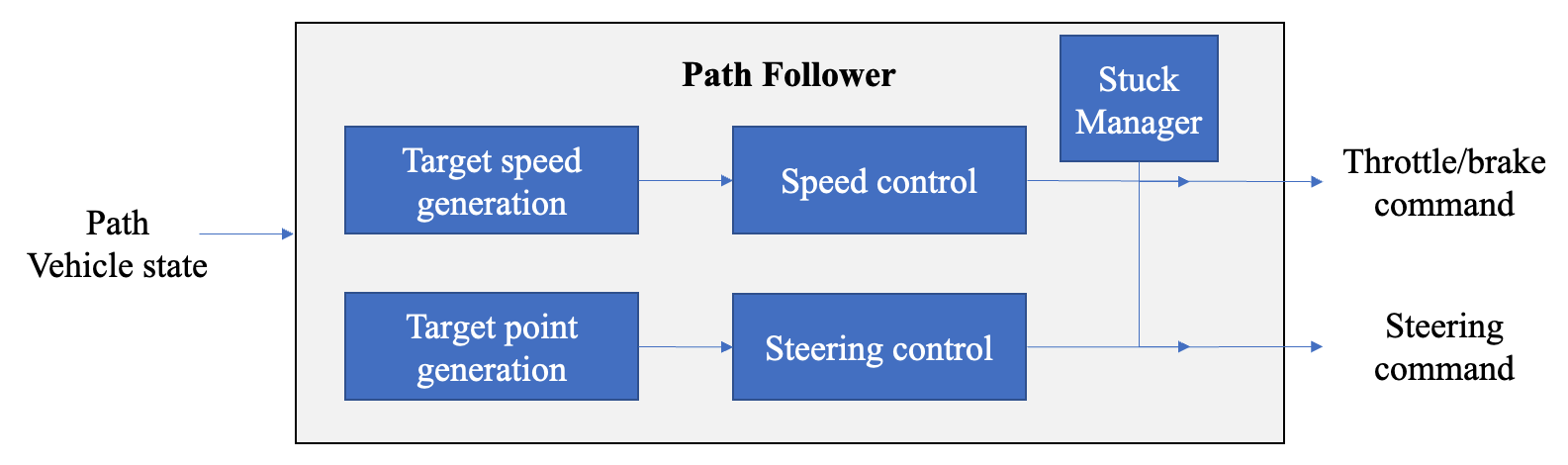}
    \caption{Core modules of the proposed path follower.}
    \label{fig:prelim2}
\end{figure}

\minisection{Method Overview.}
The path follower technique used in this work consists of five core modules: target speed generation, target point generation, speed control, steering control, and stuck manager as shown in Figure \ref{fig:prelim2}. Our core contribution is the target speed generation module which uses a novel algorithm for efficiently calculating a desired vehicle speed. For the remaining modules, we provide brief descriptions of the established approaches we apply. 

The speed controller leverages a \emph{Proportional-Integral} (PI) controller which regulates vehicle speed via throttle commands using the difference between the target speed and the vehicle's current speed, referred to as the error signal~\cite{pid1}. The PI controller uses both the proportional and integral terms of the error signal to generate a vehicle command. However, certain strategies must be implemented to mitigate common issues associated with the integral term including integral windup~\cite{pid1}. To generate a steering command, our first step is to compute a target position on the path some lookahead distance in front of the vehicle. From here, we use the geometric steering controller \emph{pure pursuit}~\cite{wallace1985first} which is based on a steering angle control law that minimizes the cross track error characterized by the target position. This algorithm effectively computes a gain at each time step for a proportional feedback controller using geometric properties of the path tracking problem in the case of a bicycle kinematic model. In case of a \emph{stuck event} being triggered, which we define as a situation where the vehicle fails to make sufficient progress along the path within a certain time frame, the commands outputted from the speed and steering controllers are overridden by the stuck manager module. For instance, we invert the sign of the throttle input during several frames in the hopes that we recover to a position where the vehicle is no longer stuck. Another potential strategy is to simply teleport the vehicle to a new position. For further strategies, we refer readers to existing works \cite{onieva2012evolutionary}.

\minisection{Target Speed Generation.}
\label{sec:target_speed}
The high-level idea of our proposed method is to limit vehicle speed based on the estimated maximum curvature of the part of the path that extends from the vehicle position to a user-defined lookahead distance. The cheap computational cost of our approach is underpinned by an analytical expression for computing the maximum curvature of a quadratic B\'ezier curve. We first describe the inversely proportional relationship between target speed and path curvature.

A commonly used method for computing a target speed is to determine the optimal speed for a vehicle to travel through a turn of a given radius while maintaining tire traction. Often referred as \emph{critical speed}, the following equation relates a vehicle's safe turning speed to the balance between centripetal force and maximum lateral force that the tires can handle~\cite{gillespie2021fundamentals}:
\begin{equation}
    v_{\text{target}} = \sqrt{\frac{a_{\text{lat}}\cdot g}{\kappa}},
\end{equation}
where $a_{\text{lat}} > 0$ is a tuning parameter, $g$ is gravitational acceleration, and $\kappa$ is curvature. The parameter $a_{\text{lat}}$ allows us to effectively tune vehicle speed around corners based on the handling of a particular vehicle. As a point of reference, the maximum lateral acceleration found in human normal driving is $0.4g$ \cite{bosetti2014human}. Once a suitable value has been specified for $a_{\text{lat}}$, the problem of computing target speed reduces to finding curvature $\kappa$. In a typical path following setting, this refers to the curvature at a particular point in the path. However, this assumes that curvature information of the path is readily available, e.g. the case of a path described by a function that is twice differentiable. If the path is defined by a list of waypoints, then the approach used here is to use curve-fitting functions to estimate the path curvature, which provides a candidate representation of the path curvature profile.

In our approach, we use a series of quadratic B\'ezier curves to efficiently capture an estimated curvature profile of the path $P$. The following is the parametric equation $\textbf{B}(t)$ which defines a quadratic B\'ezier curve with control points $\textbf{p}_1, \textbf{p}_2, \textbf{p}_3$:
\begin{equation}
    \textbf{B}(t) = (1-t)^2\textbf{p}_1 + 2(1-t)t\textbf{p}_2 + t^2\textbf{p}_3, \ 0 \leq t \leq 1.
\end{equation}
The first and third control points $\textbf{p}_1, \textbf{p}_3$ are the endpoints (at $t=0$ and $t=1$, respectively) while $\textbf{p}_2$ generally does not lie on the curve. To compute curvature along the curve, we may use the standard curvature equation $\kappa(t) = \lvert \textbf{B}(t)'\times \textbf{B}(t)''\rvert\lVert \textbf{B}(t)' \rVert^{-3}$, and a sampling strategy to sample curvature of points along the curve. However, a more efficient approach is to leverage the particular geometric constraints of quadratic B\'ezier curves. Specifically, the maximum curvature can be computed analytically, allowing us to determine an upper bound for the curvature profile without the need for sampling. This is the core motivation for employing quadratic B\'ezier curves over higher order curves which may provide better estimates of the curvature profile though at a much higher computation cost.

\begin{figure}[!t]
    \centering
    \includegraphics[width=0.75\columnwidth]{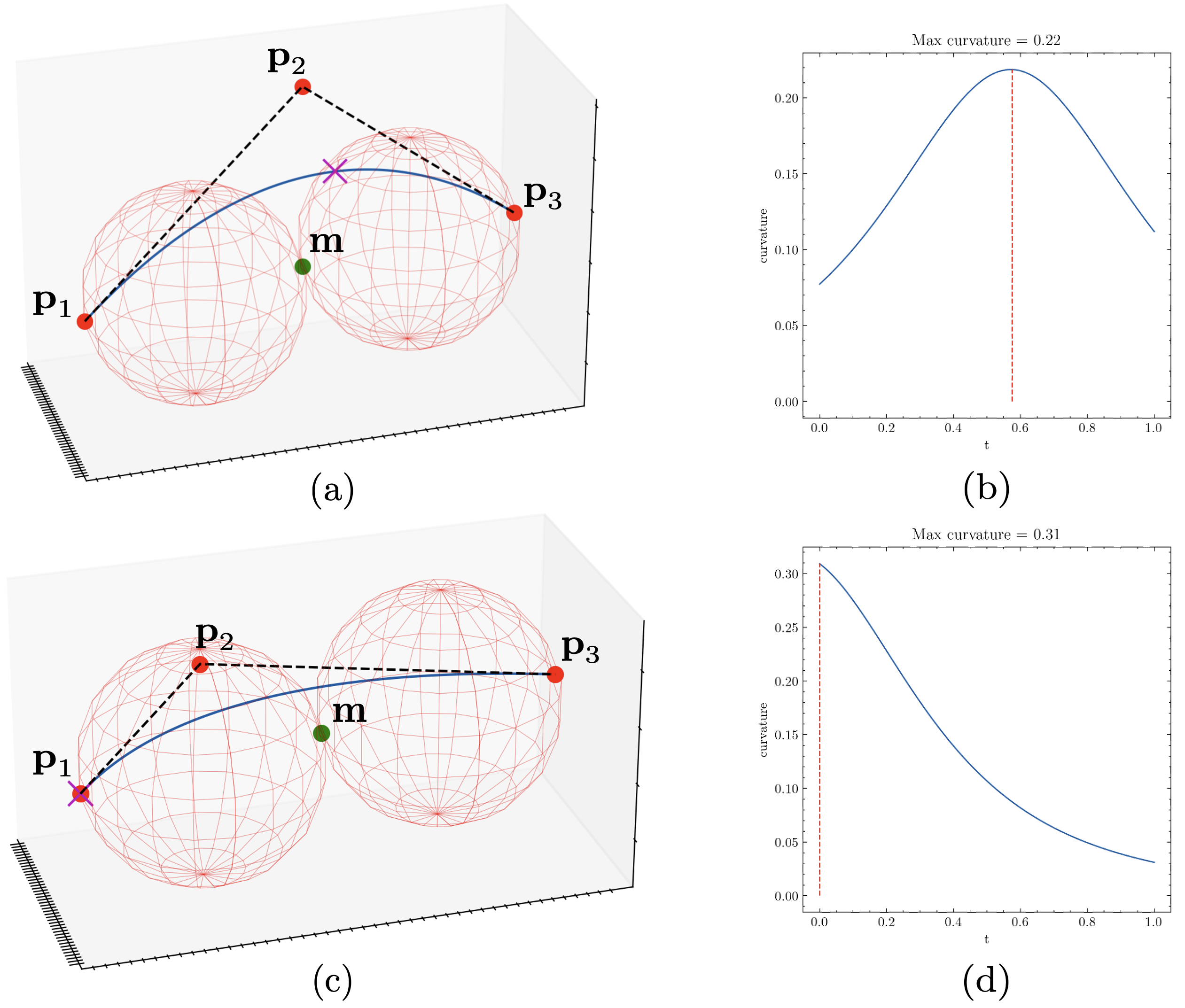}
    \caption{Sample quadratic B\'ezier curves (blue) demonstrating the two cases for the curvature profile. In the first case (a), the middle control point $\textbf{p}_2$ lies outside the spheres, producing a unimodal curvature function (b). In the second case (c), $\textbf{p}_2$ is inside one of the spheres, resulting in a curvature function that is monotone (d).}
    \label{fig:quadBezier}
\end{figure}

Figure \ref{fig:quadBezier}(a) shows a sample quadratic B\'ezier curve and we highlight the two equally sized spheres drawn along the segment $\textbf{p}_1\textbf{p}_3$ which meet at the midpoint $\textbf{m}$. The radius of these spheres is equal to half the euclidean distance between the endpoint and the midpoint $\textbf{m}$, i.e. $r = \frac{1}{2}\lVert\textbf{p}_1 - \textbf{m}\rVert$.
There are two possible cases that characterize the curvature profile of $\textbf{B}(t)$ based on where the control point $\textbf{p}_2$ lies in relation to these two spheres. If $\textbf{p}_2$ lies outside of the two spheres, i.e. $\lVert \textbf{p}_2 - \frac{1}{2}(\textbf{p}_1 + \textbf{m}) \rVert > r$ and $\lVert \textbf{p}_2 - \frac{1}{2}(\textbf{p}_3 + \textbf{m}) \rVert > r$, then from the work by \citet{deddi2000interpolation} the maximum curvature of $\textbf{B}(t)$ is given by the expression:
\begin{equation}
    \kappa_{\max} = \frac{{\lVert \textbf{p}_2 - \textbf{m} \rVert}^3}{\left(\frac{1}{2}\lVert (\textbf{p}_1 - \textbf{p}_2) \times (\textbf{p}_1 - \textbf{p}_3\right) \rVert)^2}.
    \label{eq:speed1}
\end{equation}
Note, the denominator is the squared area $A$ of the triangle characterized by the three control points. In the second case, $\textbf{p}_2$ lies inside one of the two spheres which means the curvature of $\textbf{B}(t)$ is monotone \cite{sapidis1992controlling}. This implies the maximum curvature occurs at either of the two endpoints $\textbf{B}(0) = \textbf{p}_1$, $\textbf{B}(1) = \textbf{p}_3$ and therefore maximum curvature is the expression \cite{deddi2000interpolation}:

\begin{equation}
\begin{split}
\kappa_{\max} = & \max(\kappa_1,  \kappa_2), \\
\kappa_1 = \frac{A}{\lVert\textbf{p}_1 - \textbf{p}_2\rVert^3}, & \;\;\;
\kappa_2 = \frac{A}{\lVert\textbf{p}_3 - \textbf{p}_2\rVert^3}.
\end{split}
\label{eq:speed2}
\end{equation}

With these two results, we can efficiently compute the maximum curvature of an arbitrary quadratic B\'ezier curve without having to sample points along the curve like in the case of more complex parametric curves.

Algorithm \ref{alg:targetSpeed} summarizes our approach to computing a target speed $v_{\text{target}}$ given inputs vehicle position $\textbf{r}$, the path to be followed $P$, and parameters maximum lateral acceleration $a_{\text{lat}}$, waypoint spacing $\Delta h$, waypoint count $N$, and speed limits $v_{\min}$, $v_{\max}$. The first major step involves constructing a new path $\Tilde{P}$ by sampling points from the original path $P$ at regular intervals of size $\Delta h$ (Lines 1-4). By capturing the salient features of the original path structure up to a lookahead distance $\Delta h \cdot N$ from the vehicle's current position $\textbf{r}$, we are able to determine appropriate target speeds in advance. This is particularly useful in areas where the vehicle needs to slow down, such as sharp corners. The purpose of initializing $\Tilde{P}$ with $\textbf{r}$ (Line 1) is made more clear in the steps to follow. In the next step, we iterate through the waypoints of our new path $\Tilde{P}$, and calculate the maximum curvature $\kappa_{\max}$ among all the quadratic B\'ezier curves that can be created from three sequential waypoints in $\Tilde{P}$ (Lines 5-9). We use Equation \eqref{eq:speed1} and \eqref{eq:speed2} to compute the maximum curvature of each curve. Since the first B\'ezier curve is always defined by the vehicle position and two points on the path $P$, we ensure the target speed is decreased whenever the vehicle is too far away from the path. In the final step of our algorithm, we use Equation \eqref{alg:targetSpeed} to compute target speed given $\kappa_{\max}$ and $a_{\text{lat}}$ followed by clamping on $v_{\text{target}}$ given limits $v_{\min}$, $v_{\max}$.

\begin{algorithm}[!h]
\scriptsize
\caption{Compute Target Speed}\label{alg:targetSpeed}
 \hspace*{\algorithmicindent} \textbf{Input:} $\textbf{r}$, $P$, $a_{\text{lat}}$, $\Delta h$, $N$, $v_{\min}$, $v_{\max}$
\begin{algorithmic}[1]
\State Initialize empty array $\Tilde{P}$ with $\textbf{r}$
\For{$i = 2,3,...,N$}
    \State Compute next point $\textbf{p}_{i}$ on path $P$ a distance $\Delta h$ away from $\Tilde{P}[i-1]$
    \State Append $\textbf{p}_{i}$ to $\Tilde{P}$
\EndFor
\State $\kappa_{\max} \gets 0$
\For{$i = 1,2,...,N-2$}
    \State $c_i \gets \Tilde{P}[i], \Tilde{P}[i+1], \Tilde{P}[i+2]$
    \State Compute max curvature $\kappa_i$ for B\'ezier curve with control points $c_i$
    \State  $\kappa_{\max} \gets \max(\kappa_{\max}, \kappa_i)$ 
\EndFor
\State $v_{\text{target}} \gets\sqrt{(a_{\text{lat}}g)(\kappa_{\max})^{-1}}$
\State $v_{\text{target}} \gets \text{clamp}(v_{\text{target}}, v_{\min}, v_{\max})$
\State \textbf{return} $v_{\text{target}}$
\end{algorithmic}
\end{algorithm}

\section{Experimental Setup}
\label{sec:exp_setup}
In this section, we describe the experimental setup used for collecting results for validating our proposed path follower. We used a first-person shooter game (Battlefield 2042) as the platform for running experiments and used a machine with an NVIDIA Tesla T4 GPU with a 16-core CPU and 110 GB of RAM. The game contains a wide selection of ground vehicles with varying driving characteristics. We integrated the proposed path follower with the test automation system which deploys AI bots for testing large multiplayer games, referred to as AutoPlayers \cite{autoplayers1}. To compare our approach, we used as a baseline the original path following logic present in AutoPlayers. Briefly, the path following logic leverages a series of heuristics that fail to generalize well across certain vehicles and environments. For instance, the target speed module computes a target inversely proportional to the largest angle between the vehicle forward vector and a segment on the path in front of the vehicle up until a lookahead distance. Such an approach can lead to noisy target speed outputs when the path is complex or contains sharp turns. In addition, the speed controller employs the simpler P-controller (i.e. no integral term) meaning no history of previous errors is kept to influence the throttle command.

Several metrics were used to assess the performance of the proposed path following solutions. During each frame, we recorded the cross track error, and indicators whether the vehicle is inside the path corridor and whether the vehicle is stuck. We consider the total number of stuck events to be the core metric of interest since its result has the widest set of implications. In the context of bots for testing, we would like to minimize the number of times vehicles get stuck to ensure good quality performance tests and effective soak tests. If vehicles get stuck too often, then the resulting test data may provide a poor representation of real gameplay and may not reflect realistic game scenarios. We note that it is very difficult to completely avoid vehicles getting stuck since the generated paths do not account for the vehicle kinematics and dynamics.

The following parameters were set for the target speed generation module: $a_{\text{lat}}=0.4$, $\Delta h=6 \ m$, $N = 5$, $v_{\min}=1 \ m/s$, $v_{\max} = 10 \ m/s$. For the parameters that are the same across the two approaches (e.g. min/max target speed), the same values were set to ensure fair comparisons.

\section{Results}
In our first set of experiments, we assessed the ability of different vehicles to follow a set of predefined \emph{test paths}. For each test path, a particular vehicle is spawned at the first waypoint and then commanded to follow the path until the final waypoint. We generated 10 test paths capturing a wide range of challenging environments such as steep hills, narrow corridors, and sharp turns. Further, six different vehicles, each with a distinct set of driving characteristics, were considered for evaluation. These vehicles include Storm, Bolte, Panhard Crab, Zaha, Armata, and Hovercraft. The first three vehicles were deemed easily maneuverable, while the latter three presented greater challenges due to slower turning rate, larger turning radius, poor traction, etc. We performed a total of $60$ tests, $10$ for each one of the $6$ vehicles.

Table \ref{tab:results1} summarizes the performance of three path followers: the baseline, our proposed solution with parameters from Section \ref{sec:exp_setup}, and our proposed solution with per-vehicle-tuned parameters including maximum lateral acceleration parameter $a_{\text{lat}}$ from the target speed generation module. The reported metrics include the number of trials with at least one stuck event, the total number of stuck events as well as a breakdown per vehicle, mean cross track error, percentage of time spent inside the path corridor, total time taken to complete the path, and mean vehicle speed. The first major result is the $70\%$ decrease in the total number of stucks events when comparing our approach to the baseline. In the case of the hovercraft, the addition of the integral term in the speed controller helped resolve the issue of getting stuck while driving up steep hills. We see similar path following improvements across the other vehicles and the smaller range of stuck events suggests our proposed solution generalizes better to different vehicle types. Our approach with fixed parameters maintained similar performance as the baseline for tracking error and inside corridor percentage while driving vehicles at faster speeds. We improve in these two metrics when we use our approach with per-vehicle tuned parameters. With faster average vehicle speeds and a further decrease in number of stuck events, we achieved a $39\%$ decrease in mean total time when comparing with the baseline. 

\begin{table}[h]
\centering
\scalebox{0.8}
{
\begin{tabular}{|l|c|c|c|}
\hline
 & Baseline & Ours & Ours (vehicle-tuned)\\
\hline
Total stuck trials ($/60$) $\downarrow$ & 33 & 15 & \textbf{8} \\
\hline
Total stuck events $\downarrow$ & 56 & 17 & $\textbf{11}$ \\
\hline
Cross track error mean (m)  $\downarrow$ & 1.46 & 1.48 & \textbf{1.32}\\
\hline
Total time mean (s) $\downarrow$ & 59.6 & 40.9 & \textbf{36.1} \\
\hline
Inside corridor mean (\%) $\uparrow$ & 91 & 91 & \textbf{93}\\
\hline
Speed mean (m/s) $\uparrow$ & 4.55 & 5.82 & \textbf{6.84}\\
\hline
& \multicolumn{3}{|c|}{Breakdown of stuck events by vehicle $\downarrow$} \\
\hline
Storm & 3 & 1 & \textbf{1} \\
Bolte & 9 & 2 & \textbf{0} \\
Panhard Crab & 8 & 4 & \textbf{2} \\
Zaha & 5 & 3 & \textbf{2} \\
Armata & 11 & 5 & \textbf{5} \\
Hovercraft & 20 & 2 & \textbf{1} \\
\hline
\end{tabular}
}
\caption{Summary of the results. The up arrow means higher is better. The down arrow means lower is better. Increased mean vehicle speed is better unless other metrics degrade.}
\label{tab:results1}
\end{table}

To assess the performance of our proposed method in a more realistic game scenario, we performed soak tests. These tests consisted of 5-minute game sessions in conquest mode and a specific map with 64 AI-controlled bots divided into two teams. A total of 5 randomized trials were ran for both the baseline and proposed path follower. We normalized the number of stuck events by the total number of seconds spent driving a ground vehicle across all 64 players. For the proposed approach, a ground vehicle was stuck on average every $131 \text{s}$ ($\sigma=36$) of driving time and every $60 \text{s}$ ($\sigma=7$) with the baseline. This means, we were more than twice as likely to get stuck with the baseline path follower compared to our proposed solution. 

\section{Conclusions and Future Work}
In this short paper we tackled the path following problem in the context of game AI. Our focus was on developing a simple and efficient framework that achieves reasonable performance across ground vehicles and game environments. Such a framework is particularly beneficial for bots that test games in development or for developing a benchmark solution. The main contribution of this work is the novel target speed generation algorithm which uses quadratic B\'ezier curves to efficiently estimate the curvature profile of a path. 
In future work, we plan to expand our framework to include aerial vehicles and to test the performance of our approach with a larger collection of ground vehicle types and game environments. Additionally, we intend to explore the use of automatic tuning methods such as black box optimization which may significantly increase path following performance and eliminate manual parameter tuning.

{\footnotesize \bibliography{bib}}
\bibliographystyle{IEEEtranN}

\end{document}